\documentclass[letterpaper, 10 pt, conference]{IEEEtran} % use above line letter sized paper 
\IEEEoverridecommandlockouts % Needed if you want to use the \thanks command 
\usepackage[utf8]{inputenc} % allow utf-8 input
\usepackage[T1]{fontenc}    % use 8-bit T1 fonts
\usepackage{url}            % simple URL typesetting
\usepackage{booktabs}       % professional-quality tables
\usepackage{amsfonts}       % blackboard math symbols
\usepackage{nicefrac}       % compact symbols for 1/2, etc.
\usepackage{microtype}      % microtypography
\usepackage{graphicx}
\usepackage{dirtytalk}
\usepackage[export]{adjustbox}
\usepackage[table,x11names]{xcolor}

\ifCLASSOPTIONcompsoc
  \usepackage[nocompress]{cite}
\else
  \usepackage{cite}
\fi
\ifCLASSINFOpdf
\else
\fi
\begin{document}
\title{Identifying Spatial Relations in Images using Convolutional Neural Networks}

\author{\IEEEauthorblockN{Mandar Haldekar, Ashwinkumar Ganesan}
\IEEEauthorblockA{Dept. Of Computer Science \& Engineering,\\
UMBC,\\
Baltimore, MD\\
mandarh1, gashwin1@umbc.edu}
\and
\IEEEauthorblockN{Tim Oates}
\IEEEauthorblockA{Dept. Of Computer Science \& Engineering,\\
UMBC,\\
Baltimore, MD\\
oates@cs.umbc.edu}}


% compsoc conferences compared to traditional conferences), use this
%Telephone: (800) 555--1212, Fax: (888) 555--1212}


\maketitle

\begin{abstract}
\footnote{*The document is a preprint version.}Traditional approaches to building a large scale knowledge graph have usually relied on extracting information (entities, their properties, and relations between them) from unstructured text (e.g. Dbpedia). Recent advances in Convolutional Neural Networks (CNN) allow us to shift our focus to learning entities and relations from images, as they build robust models that require little or no pre-processing of the images. In this paper, we present an approach to identify and extract spatial relations (e.g., The girl is standing \textit{behind} the table) from images using CNNs. Our research addresses two specific challenges: providing insight into how spatial relations are learned by the network and which parts of the image are used to predict these relations. We use the pre-trained network \textit{VGGNet} to extract features from an image and train a Multi-layer Perceptron (MLP) on a set of synthetic images and the \textit{sun09} dataset to extract spatial relations. The MLP predicts spatial relations without a bounding box around the objects or the space in the image depicting the relation. To understand how the spatial relations are represented in the network, a heatmap is overlayed on the image to show the regions that are deemed \textit{important} by the network. Also, we analyze the MLP to show the relationship between the activation of consistent groups of nodes and the prediction of a spatial relation. We show how the loss of these groups affects the network's ability to identify relations.
\end{abstract}


% \fi
\IEEEpeerreviewmaketitle

\section{Introduction}
Research in the area of knowledge graphs have largely focused on mining large corpora of text to extract entities, concepts, and the relations between them. Linking entities extracted from images to knowledge graphs and reasoning with them has received little attention. Knowledge graphs (KG) such \textit{DBpedia} \cite{bizer2009dbpedia} and \textit{Yago} \cite{suchanek2007yago} store entities and their relations in the form of \textit{RDF} triples. Certain attributes contain pictures of the entity described, but most systems that use these KGs lack the ability to reason using these pictures. Pictures can provide information that can be used to improve a system's ability to perform tasks such as image search, visual verification \cite{sadeghi2015viske} (i.e., to verify the validity of a fact by looking for proof in pictures) and visual question answering \cite{antol2015vqa} (i.e. answering questions about the image). To improve visual reasoning, Zhu et. al \cite{zhu2014reasoning} use a knowledge graph representation to reason about object affordances \cite{gibson2014ecological}, where the system tries to learn about the object, estimate the human pose to perform an action over it, and define the relative position of object with respect to the human. These tasks have become important in an age where smart-phones with cameras are the preferred mode of communication and to search for content on the web \cite{WebSearch}.

Object detection, scene description, pose estimation and other such tasks were commonly solved using SIFT \cite{lowe2004distinctive} and HOG \cite{dalal2005histograms}  features. But in recent years, deep neural networks have had tremendous success in diverse fields such as computer vision \cite{NIPS2012_4824} \cite{girshick2014rcnn}, natural language processing \cite{TextCNN}\cite{Socher-etal:2013} and audio \cite{SpeechRecWithDNN}. These developments, coupled with the availability of large quantities of data and high performance computing hardware such as GPUs (with CUDA and dedicated libraries like CuDNN), made remarkable performance gains possible.  One of the popular deep neural network architectures, a Convolutional Neural Network (CNN), has substantially improved image classification compared to other existing methods as shown by Krizhevsky et. al. in 2012 \cite{NIPS2012_4824}. CNNs are also able to achieve state-of-the-art performance in many of the aforementioned tasks \cite{girshick2014rcnn}\cite{ poseactionrcnn}\cite{Zhou}.

\begin{figure}[h]
    \centering
    \includegraphics[width=0.4\textwidth]{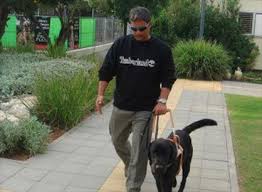}
    \caption{Person is \textit{beside} a dog}
    \label{fig:person_nextto_dog}
\end{figure}

\begin{figure*}[h]
    \centering
    \includegraphics[scale=0.46]{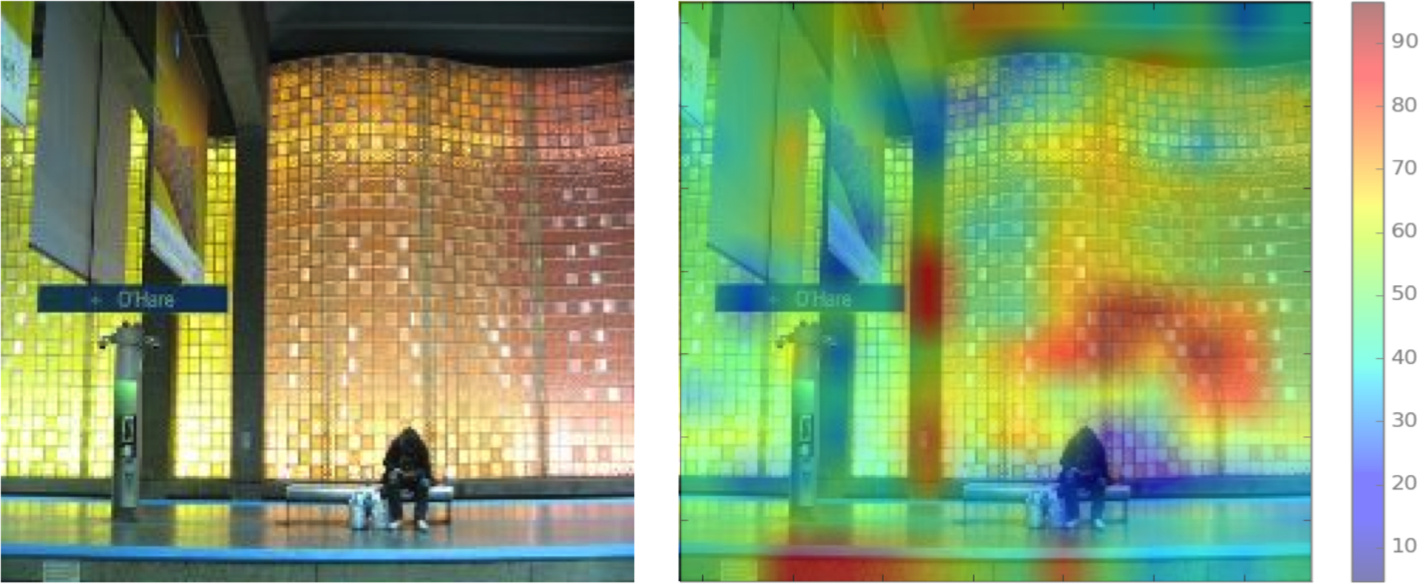}
    \caption{An image and the associated heatmap showing important parts of the image}
    \label{fig:sparelations}
\end{figure*}

However, despite their success, intermittent work has been conducted to understand how deep neural networks learn features from the input data. Our focus is to extend the understanding of how deep networks operate by looking at the network's representations of spatial relations between objects in images. Spatial relations are difficult to learn as they are defined by the relative positions of objects in the image and are independent of the object itself (e.g., the relation \textit{next to} is common to both scenarios: (a) \textit{A man is standing next to his dog} (b) \textit{A child is sitting next to the chair}). Consider figure \ref{fig:person_nextto_dog} where the desired output is to extract the relation \textit{next-to} directly from the image and represent it as:
\[\exists{x} \exists{y}: Person(x) \land Dog(y) \land Beside(x,y)\] Our research focuses on training a network to classify a subset of spatial relations only. Depending on the frame of reference, spatial relations are classified into three types: \textit{basic}, \textit{deictic}, or \textit{intrinsic} \cite{logan1996computational}. Our network is trained on \textit{deictic} relations, where the relation between two objects is specified with respect to the viewer's point of view (POV). In this paper, we show how the network learns these relations by overlaying a generated heatmap on the original image that highlights parts of the image that are considered \textit{important} to identify the relations. On further analysis, experiments reveal how a group of nodes are formed in the network which have a higher impact on the ability of the network to predict them. Figure \ref{fig:sparelations} shows an example image and the overlayed heatmap. The regions in red are important to  identify the relation \textit{behind}.

The following sections discuss the background and related work, the architecture used to train a model to predict spatial relations, the datasets on which the models are trained, and the observations from experiments conducted.

\section{Related Work}
A convolutional neural network (CNN) is a feed forward network that uses a combination of convolutional layers, pooling layers, and fully-connected layers. It is common to train with a batch of input images (determined during fine tuning of the network) and the gradient is calculated for the average loss across the batch. There are multiple CNN architectures that have been proposed to classify an object in images such AlexNet \cite{Krizhevsky2009}, VGGNet \cite{Simonyan14c} and Inception \cite{szegedy2015going}. One of the key reasons why CNNs  are effective is because the network is trained on large datasets (such as imagenet) and can be directly applied (with fine tuning) to perform object recognition on images from different domains (domain adaptation) as well as used for various visual recognition tasks \cite{DBLP:journals/corr/DonahueJVHZTD13}.

\subsection{VGGNet}
Like many CNN architectures, VGGNet \cite{Simonyan14c} is trained on the imagenet (ILSVRC) dataset \cite{deng2009imagenet}. The network uses a small receptive field of size 3x3 and 5 spatial pooling layers, and ReLU as the activation function. It has 2 fully-connected layers at the end before the output layer, which has 1000 nodes, representing all the labels in imagenet. The fully-connected layers are each of size 4096. VGGNet is known to be robust. Output of specific layers are used to describe images \cite{karpathy2015deep} (layer FC-7). Our pipeline also utilizes the output of layer FC-7. A 16 layer pre-trained VGGNet model is used in our experiments.

\subsection{Spatial Relations}
Prior research on recognizing spatial relations between objects in images relies on constructing handcrafted features by manually inspecting the dataset \cite{lan2012image}. Humans, though, perceive the relative positions of objects by fitting their positions into one of the pre-defined templates, i.e., to identify the relation \cite{logan1996computational}. Malinowski et. al \cite{mmalinowski14spatialpooling} create spatial templates for each relation and then use a bi-directional fragment embedding framework to associate the template with text triplets \cite{karpathy}. To train their model to perform spatial reasoning, they annotate the SUN09 dataset with spatial relations. For example, the image in figure \ref{fig:sparelations} is annotated as a bench (subject) \textit{in-front-of} a wall (object). In our work, we identify the spatial relation without using the bounding box surrounding the objects in the image and the associated annotation. We show that learning to identify the presence of an object, the relative positions of the objects, as well as the space between them is useful to identify the spatial relation.

\subsection{Interpreting CNN features}
With CNNs solving a large number of tasks, there is a growing interest to understand how features are represented in the network and the effect of various hyper-parameters used to train the network. Zeiler et. al \cite{Zeiler2010} create a deconvolutional network to show the mid-level representations in a convolutional neural network. They show that features learned at intermediate layers go from a lower level of abstraction in the initial layers (like basic shape features) to higher level abstraction (such as texture) in the layers ahead. Also, they occlude parts of the image to understand which parts are important to classify the objects in the image. Simoyan et al. \cite{Simonyan2013} backpropagate the gradient of the class score with respect to the image pixels to generate an artificial image which is representative of what is learned for that particular class. The method can be utilized for weakly-supervised object localization. Long et al.\cite{Long2014} prove that despite using large receptive fields, CNNs learn the correspondence between object parts. 

 Escorcia et. al. \cite{Escorcia2015} provide empirical evidence of attribute centric nodes (ACNs) inside a CNN that are trained to recognize specific objects. These ACNs represents different attributes in the image such as \textit{texture} like furry, black/brown, wooden, etc. Their experiments reveal that the network recognizes the aforementioned attributes even though it is not specifically trained for it. ACNs are sparse and their distribution across the layers is not uniform, but are likely to be found in the top (forward) layers of the network. They also show that when ACNs corresponding to a certain sets of attributes are ablated, it reduces classification accuracy of the objects represented by these attributes. 

Another way to interpret what the network is learning is by isolating which parts of the image are \textit{important} for the network to classify it. Although this can be achieved by a class activation mapping technique which finds the regions in a single forward pass of the network \cite{Zhou2015}, a fully convolutional neural network without any fully connected layers is required.

In our approach, we similarly try to isolate the parts of the image that are considered \textit{important} for the purpose of classification. We mask each region of the image and measure the entropy change to characterize the impact of each region of the image and ablate nodes to check impact on the relations predicted. The heatmap is generated even though fully connected layers are present. The network is separately trained on the SUN09 dataset and a synthetic dataset generated for the purpose of studying the network.

\section{Preliminaries \& Analysis Methods}
\subsection{Cross Entropy}
Consider a trained neural network $A$. Given an image $X^{(i)}$ of size $M$x$N$ and the spatial relation labels $y_1 ... y_z$, the softmax cross entropy loss function is:
\begin{equation}
C^{(i)} = - \sum_{y=1}^{z} P_i(y) log(Q_i(y))
\label{eq:cross_entropy}
\end{equation}
where $C^{(i)}$ is the cross entropy, $Q_i(y)$ is the softmax probability of the label and $P_i(y)$ is actual value of the label. An input image can be modified for a number of reasons such as identifying the important regions of an image, isolating a particular object in the image, etc. Let $X^{(i)}_j$ be the modified version of the original image $X^{(i)}$. Using eq. \ref{eq:cross_entropy}, the cross entropy of the new image  $C^{(i)}_j$  (with the label of the original image) can be calculated ($C^{(i)}$ can be considered to be the baseline cross entropy).

The entropy difference is given by:
\begin{equation}
E^{(i)} = C^{(i)}_j - C^{(i)}
\label{eq:entropy_diff}
\end{equation}
The entropy difference is used to measure the effect on the prediction while a network tries to classify the image when any change is made to it or to the network architecture.

\subsection{Characterizing the Cross Entropy Difference}
During training, the network reduces the loss between the predicted and true label. Once a network is trained, the magnitude of $E^{(i)}$ shows the \textit{importance} of the change to the image. The larger the magnitude of $E^{(i)}$, the higher is the likely importance. We call this the \textit{influence} of the introduced change (in the image or the network) on the ability of the network to perform a prediction. There are three possible conditions:
\begin{itemize}
    \item $E^{(i)} > 0$: \textit{influence} is positive, i.e., the change introduced increases the likelihood of the image to be identified as the label.
    \item $E^{(i)} < 0$: \textit{influence} is negative, i.e., the change introduced reduces the likelihood of the image to be identified as the label.
    \item $E^{(i)} = 0$: No \textit{influence}.
\end{itemize}

\subsection{Ablating Nodes in Layers}
Consider a fully connected layer $l_i$ of size $n$. The previous layer $l_{i-1}$ has size $h$, the resulting output being $O_{i-1}$ (size $1{\times}h$). The weight vector is $W_{h \times n}$ and $b_{1 \times n}$ is the bias. A fully connected layer generates an output:
\[O_i = max(0, W \cdot O_{i-1} + b)\]
where a Rectified Linear Unit (ReLU) is the activation function. We perform an element-wise multiplication with a vector $A_{1 \times n}$.
\[AO_i = O_i \odot A\]
A is the \textit{ablation} vector. The vector contains 1 at every index position except the nodes that need to be ablated which have a value of 0. $AO_i$ is the final output of the layer that is transferred forward to the next layer. Because we use a ReLU, the node might have a final output of 0 or a value greater than 0. Thus the above method will ablate nodes only in the case where the output is greater than 0. The method sufficiently ablates the nodes because in the case where the output of a node is 0, the node does not impact any forward computation in the proceeding layers.

\subsection{Extract Important Regions of Images}
Important regions are parts of the image that the network is paying the most attention to while classifying images for particular class labels. The key idea is to artificially mask small regions of the image and measure the \textit{influence} (eq. \ref{eq:entropy_diff}). The mask is a patch placed on the specific part of the image that makes the region unrecognizable, like a gray colored rectangular mask for images with non-gray colored backgrounds.

Below is the procedure to measure the \textit{influence}:\\
Consider a trained network $N$. Given an image $X^{(i)}$ with cross entropy $C^{(i)}$, create a gray mask for size $A$x$B$. The mask is sequentially slid on the image to create a set of images $X^{(i)}_S = \{X^{(i)}_1 ... X^{(i)}_k\}$ where each image has a single region that is masked. The number of regions $k$ depends on the size of the mask and the incremental step where the mask will be next applied on the image. The set of \textit{influence} is given by $E^{(i)}_S = \{E^{(i)}_1 ... E^{(i)}_k\}$. The region of the image with the highest \textit{influence} is:
\[r^{(i)}_{imp} = max(E^{(i)}_S)\]
represented by mask region $j$ where $X^{(i)}_j \in X^{(i)}_S$. We can define a set of regions that are influential using a threshold value $t$.

\[R^{(i)} = \{r^{(i)}_j | E^{(i)}_j > t , E^{(i)}_j \in E^{(i)}_S\}\]

In our experiments, we first mask the region present at the upper the left corner of the image and then slide the rectangular mask over the image without overlapping previous regions. We do not consider overlapping regions in order to limit the number of different images generated while masking each region of the image.

$R^{(i)}$ and the corresponding $E^{(i)}$ values are used to draw a heatmap to show the \textit{influence}. The regions that have a high positive influence are important parts of image, necessary for classification of certain spatial relations. We postulate that the network is looking for the relative position of objects as well as identifying the space between objects to classify images for spatial relations. The \textit{fineness} or \textit{coarseness} of the heatmaps can be changed by modifying the size of the mask or the iteration step used.

\subsection{Analysis of effect of nodes on spatial relations}\label{subsection:ablation}
Another way to understand how the network learns spatial relations is to analyze the internals of the architecture and isolate sets of nodes that have a positive \textit{influence} on the classification of a certain spatial relation. This is calculated by measuring the change in cross entropy when a specific node or a group of nodes are ablated (i.e. zeroed out).

Consider a trained MLP $A$ and its fully connected (FC) layer $l$. The size of the $l$ is $p$. The nodes in the layer are $N = \{n_1 ... n_p\}$ Let the relations be $Y = \{y_1 .. y_z\}$. Given an image set $X$, we first find the baseline cross entropy:\\

$C(Y) = \{C^{(i)}(y) | i \in X, y \in Y \}$\\

We ablate a single node at a time and measure change in entropy. Thus each image in the test dataset has an \textit{influence} value $E$ with respect to the ablated node. The node $j$ has an \textit{influence} corresponding to $E^{(i)(Y)}$ that is set of all the \textit{influence} values for all the relations (\textit{influence} is calculated per label). The final output is the average \textit{influence} per node per relation (label). This is to find which nodes in a layer have the highest \textit{influence} on a spatial relation. Once the nodes are identified, the clusters generated per relation are ablated simultaneously to see the net effect on the classification of a relation.

\section{System Architecture \& Datasets}
\begin{figure}[h]
    \centering
    \includegraphics[width=0.5\textwidth]{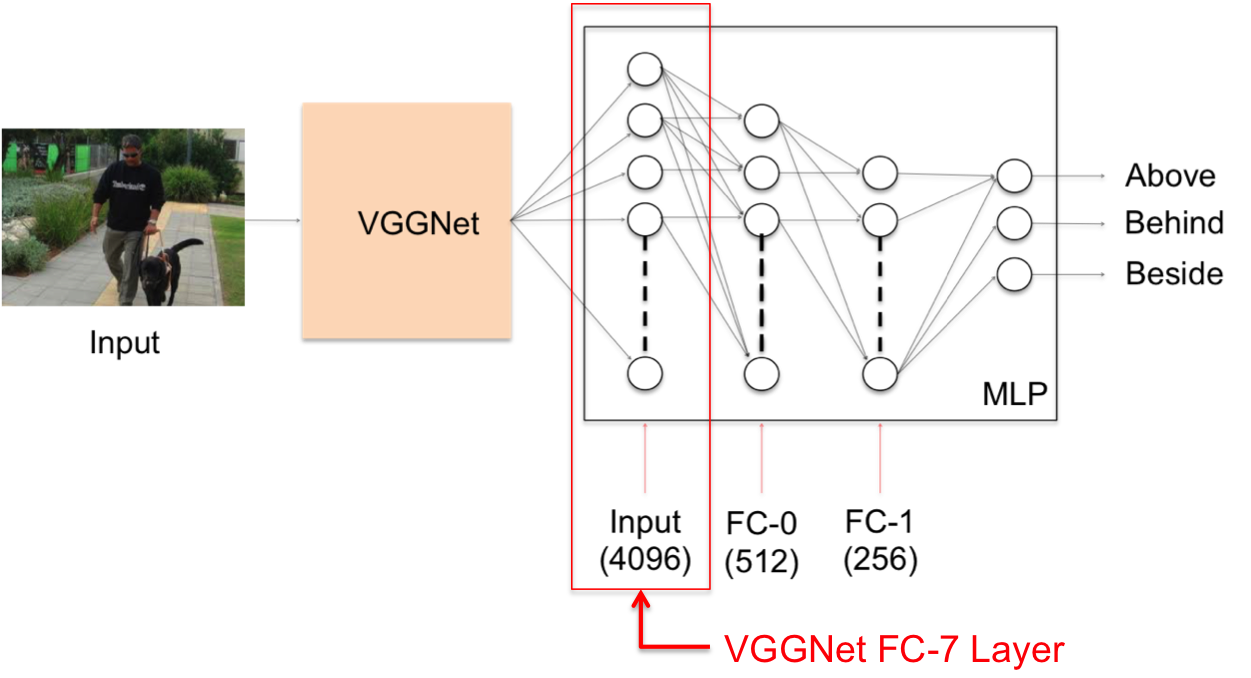}
    \caption{System pipeline: Using \textit{VGGNet} to preprocess and MLP to create a model.}
    \label{fig:VGGNET_MLP}
\end{figure}

Figure \ref{fig:VGGNET_MLP} shows the overall architecture of our system. It consists of two parts:
\begin{itemize}
    \item \textbf{Preprocessing}: The images of size 224x224 are transformed using \textit{VGGNet} to a feature vector (output of layer FC-7) that has a size of 4096. The image labels are one-hot encoded.
    \item \textbf{Model Generation}: Once the images are preprocessed, a two layer MLP is trained to detect the spatial relation. The type of relation predicted depends on the method by which the labels are encoded.
\end{itemize}

\subsection{Image Preprocessing}
 A pre-trained model of a 16-layer \textit{VGGNet} \cite{Simonyan14c} is used. It requires the input to have a standard size of $224\times224$ pixels. Hence before transforming the image, it is resized (original pictures have varying number of pixels). The process is completed in a single forward pass. While \textit{VGGNet} performs object recognition, the feature vector is extracted from layer FC-7 that is the second fully connected layer in the network. \textit{VGGNet} features rather than the images are used as input to the multi-layer perceptron (figure \ref{fig:VGGNET_MLP}) because features extracted are robust in terms of the kinds of objects detected, and are translation and rotation invariant \cite{CNNFeaturesOffTheShelf}.

\subsection{Model Generation}
The extracted image features are fed as inputs to a multi-layer perceptron as shown in figure \ref{fig:VGGNET_MLP}. The MLP contains two fully connected layers. The MLP's hidden layers are ReLU activated. The dropout rate is 0.5 and a softmax classifier is present in the output layer. Categorical cross entropy is the loss function. An AdamOptimizer calculates the gradient after each batch (of size 10). The learning rate is 0.001. \\

There are two types of datasets used to evaluate the system. A modified version of the SUN09 dataset \cite{mmalinowski14spatialpooling} and a \textit{synthetic} dataset containing a limited number of objects and relations in the image.

\subsection{SUN09 Dataset}
The original SUN09 dataset consists of around 12,000 images. Malinowski et. al. \cite{mmalinowski14spatialpooling} take a subset of them and annotate them with spatial prepositions such as \textit{above}, \textit{below}, \textit{behind} and \textit{in front of}. The images were randomly selected for annotation. The final dataset consists of 53 structured queries (i.e., tuples defined as subject-relation-object) and 11 distinct spatial relations. There are 4468 images for training and 4955 images to test.

\subsection{Synthetic Dataset}
\begin{figure}[h]
    \centering
    \includegraphics[scale=0.4]{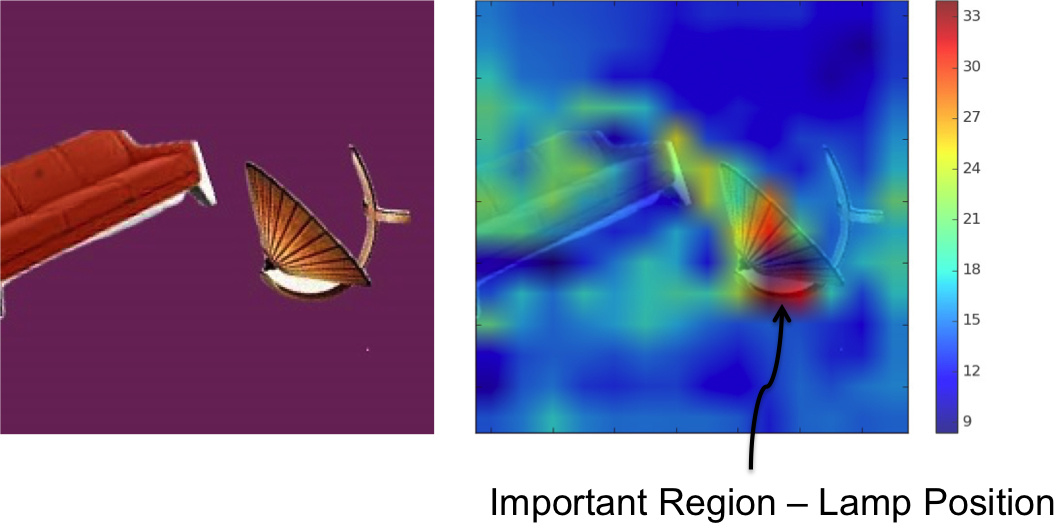}
    \caption{Sofa \textit{beside} lamp (a) Original image (b) Heat map overlayed on the image}
    \label{fig:beside1}
\end{figure}

This dataset is generated using google search on a predefined set of objects, manually selecting sample images and combining them\footnote{Python Imaging Library (PIL) is used to perform image manipulation}. It consists 10 types of objects present in the training dataset and 5 in the test dataset. The objects are:
\begin{itemize}
    \item \textbf{Training}: Dog, tiger, table, lamp, TV, sofa, ball, hat, vase and vacuum cleaner.
    \item \textbf{Test}: Deer, lion, drawer, bag, car.
\end{itemize}
Each object above has a variety of images in the dataset. For example, there are 7 different dog images and 5 different TV images. The images contain any two objects either rotated or translated to different locations in an image template with varying background colors. They are then combined to form a single image. Figure \ref{fig:beside1} shows a sample set of images.

The training dataset contains 2628 images while the test contains 432 images. Each one is annotated as object-relation-object. Background colors used in the test are completely different from those used while training. 

The advantages of the synthetic dataset are that there is a pair-wise object constraint while generating images. All objects annotated are in the foreground and distinct while the SUN09 dataset has many annotations that are between non-significant objects (e.g. figure \ref{fig:sun09_dataset}(c)).

\section{Experimental Results \& Discussion}

% Details about network size, training parameters
% what is the resulting accuracy, both test and training
%%%% BESIDE%%%%%%%%%%%%%%%

\subsection{Extracting Regions}
Table \ref{table:training_and_test_acc} shows the accuracy of the network on training as well as test data. We use images from training data (synthetic dataset) that are correctly classified by the network for our analysis.

\begin{table}[h]
\centering
\begin{tabular}{||l|l|l||}
\cline{2-3}
\multicolumn{1}{c|}{} & \cellcolor{lightgray} \textbf{Synthetic Data} & \cellcolor{lightgray} \textbf{SUN09}\\ \hline
Training Accuracy & 100\% & 71.97\%             \\ \hline
Test Accuracy     & 68.98\% & 55.98\%            \\ \hline
\end{tabular}
\vspace{2ex}
\caption{Training and test accuracy for MLP}
\label{table:training_and_test_acc}
\end{table}

We analyzed 50 training images for each of the 3 spatial relations classes that are correctly classified by the MLP.  The objects used for testing are completely separate from the objects in the synthetic dataset used for training. A gray rectangular mask of size 16 x 16 pixels covers each section of the image iteratively. Since they have been re-sized to 224 x 224 pixels, a total of 196 images are generated, once each region is masked.

\begin{figure*}[t]
    \centering
    \includegraphics[scale=0.6]{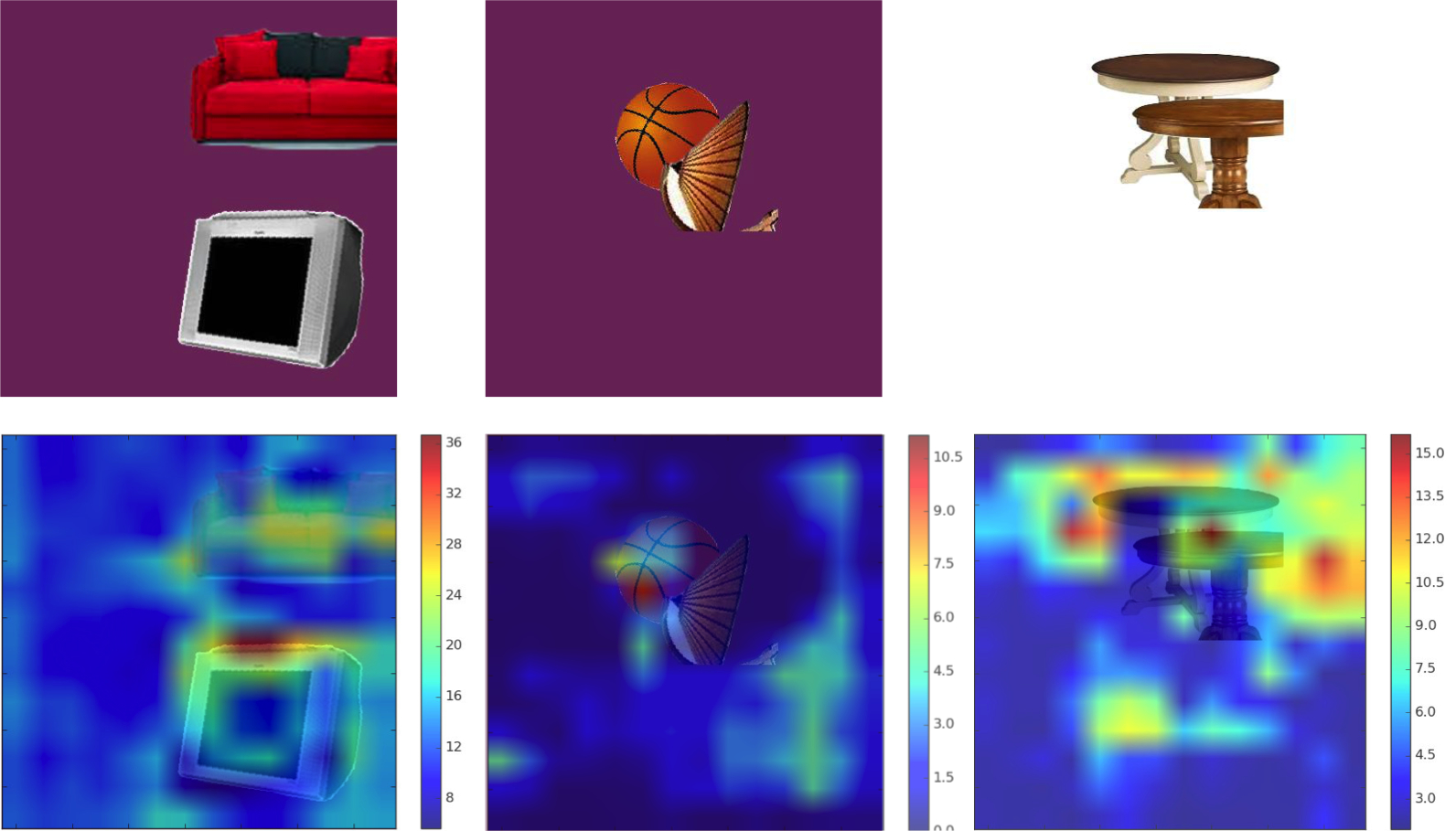}
    \caption{Different synthetic data images with their overlayed heatmaps.
             (a) Sofa \textit{above} TV (b) Ball \textit{behind} Lamp (c) Table \textit{behind} Table}
    \label{fig:toy_dataset}
\end{figure*}

Figures \ref{fig:beside1} and \ref{fig:toy_dataset} show the heatmap for some of the synthetic images. The scale of heatmaps differs from image to image. The red regions are the most important, progressively going to blue that are the least. The range of the scale is specific to an image because it is decided based on the maximum and minimum value of $E^{(i)}$ for that image.

Consider the image from figure \ref{fig:beside1} where a sofa is beside a lamp. The heat map shows that the regions of the image where the lamp is located has a large increase in cross entropy resulting in a positive $E$. Thus we see that the network is looking for the relative position of the object while classifying an image as \textit{beside}. The heatmap generated is consistent (i.e the position of the object is tracked) even when the object is rotated or moved to a separate location in the image. The network is tracking the relative position of other objects with respect to a fixed point (shown in red).

Now we look at what is being learned by the network in case of the \textit{above} spatial relation. Refer to the image in figure \ref{fig:toy_dataset}(a) where a TV is above the sofa. It is evident from the heat map that the network is looking for the relative positions of the TV as well as the space between the TV and the sofa. In addition to this, the network also learns to see gaps present between two objects, but this is not always the case.

In case of the \textit{behind} spatial relation (figure \ref{fig:toy_dataset} (b)), the corresponding heat map shows a high $E$ at the places where the ball is present and that there is a single important location in the image concentrated between the ball and lamp. Thus in case of the \say{behind} spatial relation, the network looks at the position of the object while ignoring the rest the image.

\begin{figure*}[t]
    \centering
    \includegraphics[scale=0.8]{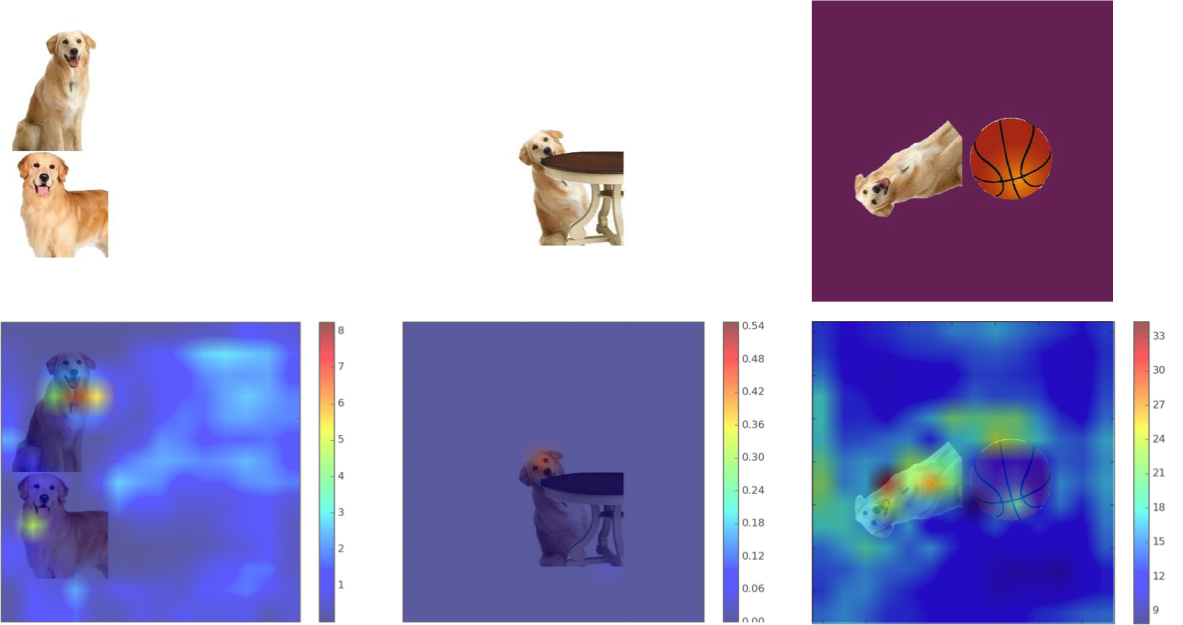}
    \caption{Images of dogs in different positions and orientations for each relation above, behind \& beside}
    \label{fig:dog_trans_rot}
\end{figure*}

\begin{figure*}[t]
    \centering
    \includegraphics[scale=0.6]{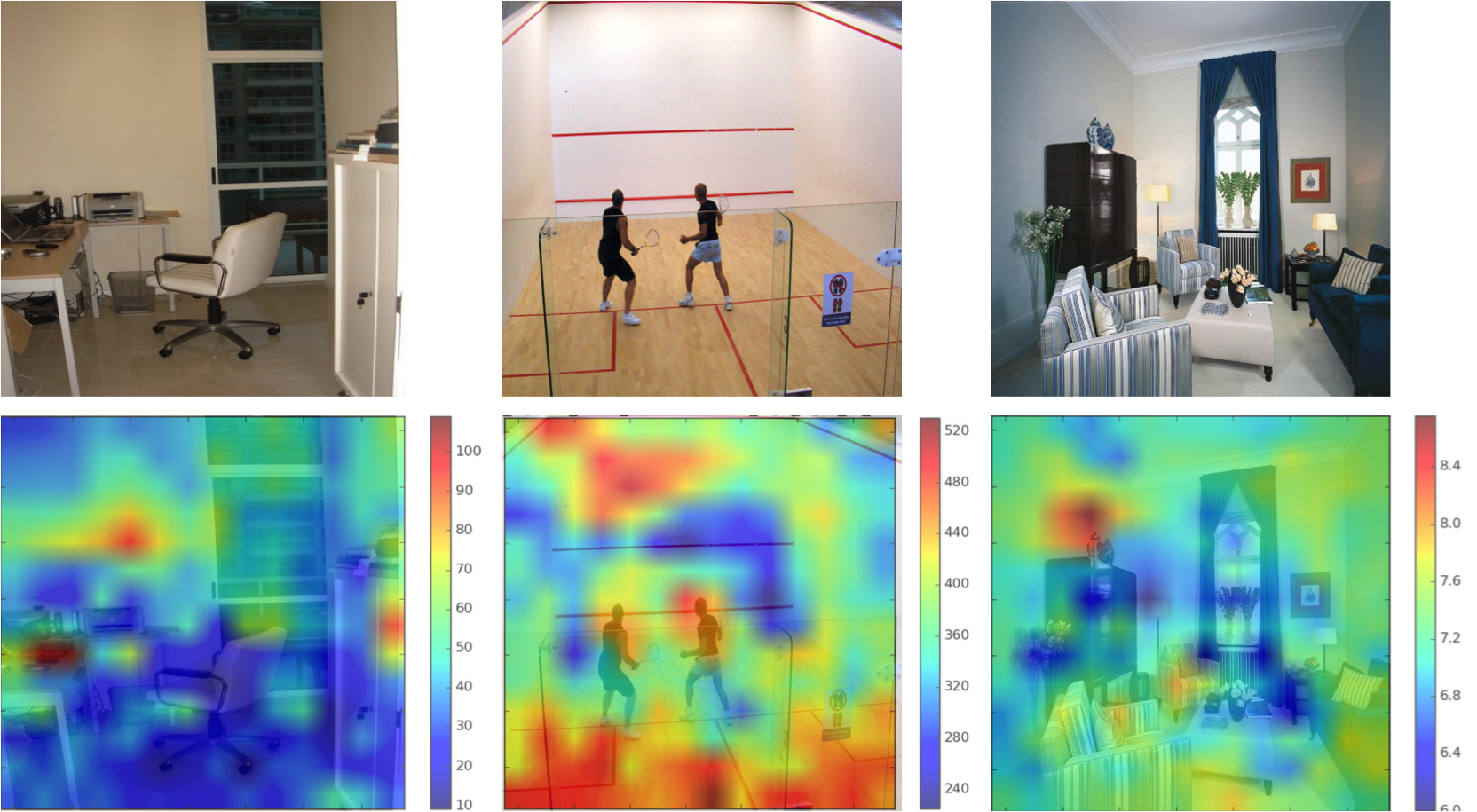}
\caption{Different SUN09 dataset images with their overlayed heatmaps. (a) Balcony \textit{in} Building (b) In squash court, Sign \textit{is right} Wall (c) Window \textit{in} Wall}
    \label{fig:sun09_dataset}
\end{figure*}

The heatmaps reveal how the relation is learned by the network. The network is trained with each object placed in different positions in the image and rotated. Figure \ref{fig:dog_trans_rot} is an example of the same object (a dog) in different positions and orientations classified correctly. The training and test accuracy in table \ref{table:training_and_test_acc} show that the network is able to look at relative positions of objects and remains unaffected by translation / rotation of objects. Figure \ref{fig:sun09_dataset} shows heatmaps for sample SUN09 images.

\subsection{Measuring Influence of Layer Nodes}
\begin{figure}[h]
    \centering
    \includegraphics[scale=0.2]{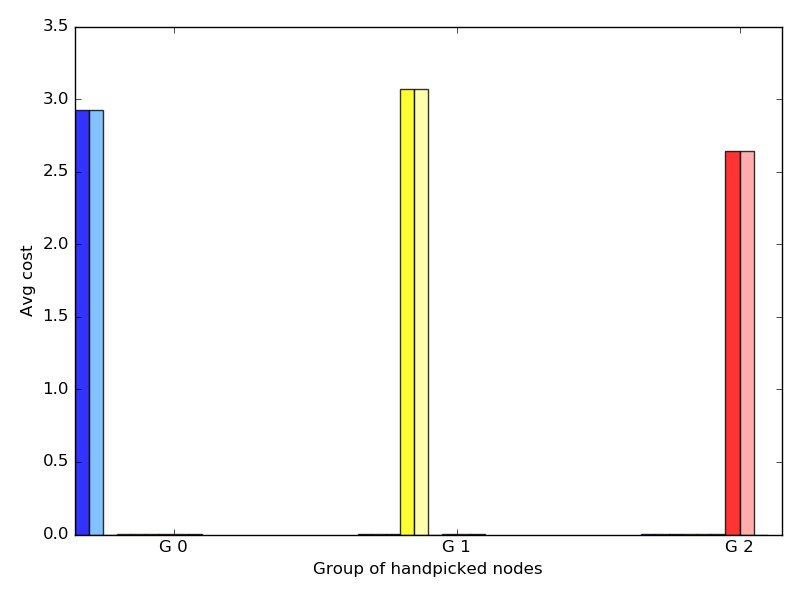}
    \includegraphics[scale=0.2]{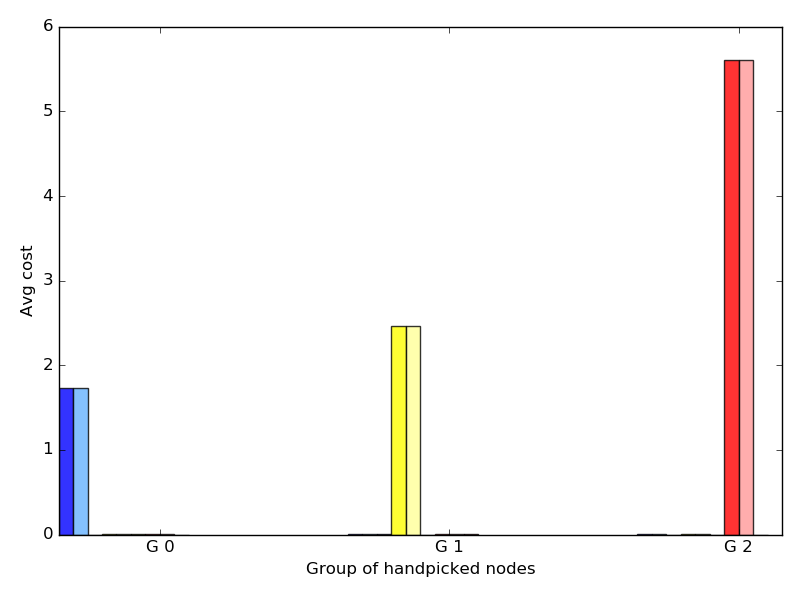}
    \caption{Average cost difference vs group of nodes zeroed out (a) Layer FC-0 (b) Layer FC-1.}
    \label{fig:handpicked_k3_layer1}
\end{figure}

We apply the method discussed in section \ref{subsection:ablation}. For this experiment, we have analyzed the training dataset (rather than test) as there is a likelihood to extract groups of nodes to avoid negative \textit{influence} from the images where the network misclassified.
For each class label, we consider the top 25\% of nodes out of the total size of the hidden layer such that zeroing them out will result in maximum increase in the cross entropy. We group these into three groups, one per class label. In this way, we get three groups of nodes where each group of nodes affects one particular spatial relation more than others. Figures \ref{fig:handpicked_k3_layer1}(a) and (b) show the differences in cross entropy vs the group that is being zeroed for the layer FC-0 and FC-1, respectively. Each group has 3 bars, i.e., absolute cost difference, positive cost difference and negative cost difference (made positive), so that each change can be characterized. The graphs show that the groups have little average negative cost (very small bars), confirming their \textit{influence} is net positive. Tables \ref{table:classwise_accuracy_layer1_handpicked} and \ref{table:classwise_accuracy_layer2_handpicked} show the accuracy with respect to each relation when these groups of nodes are ablated. For example group 0 has an average $E$ of $0.29$. The entropy change results in a reduction in accuracy of $85.7\%$ because the nodes are not utilized while classifying an image with the label \textit{Above}. The same is seen for group 1 \& 2 where the accuracy of the network to predict \textit{Beside} \& \textit{Behind} reduces to $8.4\%$ and $20.43\%$. The classification accuracy for the other labels remains $100\%$. Thus we can clearly see the existence of these sets of nodes that affect just one particular class label. The existence of such sets of nodes is more evident in layer FC-0 rather than layer FC-1. We do not see the existence of such sets of nodes clearly in the last fully connected layer. One possible reason is that the number of nodes in FC-1 are fewer (256 as compared to 512 in FC-0), so the number of nodes that are common in the three groups of nodes is more.

\begin{table}[h]
\centering
\begin{tabular}{||l|l|l|l||}
\cline{2-4}
\multicolumn{1}{c|}{} & \cellcolor{lightgray} \textbf{Above}        & \cellcolor{lightgray} \textbf{Beside}       & \cellcolor{lightgray} \textbf{Behind}       \\ \hline
Baseline Accuracy & 1.0          & 1.0          & 1.0          \\ \hline
Group 0           & \cellcolor{blue!25}0.1438       & 1.0          & 1.0       \\ \hline
Group 1           & 1.0       & \cellcolor{blue!25}0.0844       & 1.0       \\ \hline
Group 2           & 1.0       & 1.0          & \cellcolor{blue!25}0.2043      \\ \hline
\end{tabular}
\vspace{2ex}
\caption{Classwise accuracy vs group of nodes that is zeroed out in layer FC-0 of the MLP}
\label{table:classwise_accuracy_layer1_handpicked}
\end{table}

\begin{table}[h]
\centering
\begin{tabular}{||l|l|l|l||}
\cline{2-4}
\multicolumn{1}{c|}{} & \cellcolor{lightgray} \textbf{Above}        & \cellcolor{lightgray} \textbf{Beside}       & \cellcolor{lightgray} \textbf{Behind}       \\ \hline
Baseline Accuracy & 1.0          & 1.0          & 1.0          \\ \hline
Group 0           & \cellcolor{blue!25}0.05593       & 1.0       & 1.0          \\ \hline
Group 1           & 1.0       & \cellcolor{blue!25}0.0       & 1.0          \\ \hline
Group 2           & \cellcolor{blue!25}0.0022       & 1.0       & 1.0       \\ \hline
\end{tabular}
\vspace{2ex}
\caption{Classwise accuracy vs group of nodes that is zeroed out in layer FC-1 of the MLP}
\label{table:classwise_accuracy_layer2_handpicked}
\end{table}

%%%%%%%%%%% ABOVE%%%%%%%%%%%%%%%%%%%%%%%%%%%%%%%

%%%%%%%%%%%%%% BEHIND%%%%%%%%%%%%%%%

\section{Conclusions \& Future Work}
We have analyzed how an MLP (with a pre-trained CNN) learns spatial relations between objects in an image. We extract image features using CNNs and train a MLP in a supervised manner for spatial relation classification on two datasets: SUN09 and a simplified synthetic dataset. We show that the network pays attention to a certain parts of images while classifying them for spatial relations and that the network looks for relative positions of objects to identify the relation.

We also show the existence of sets of nodes representing one particular relation by conducting extensive tests to isolate single nodes that \textit{influence} a certain relation and how the group behaves when ablated. 
In future, we would like to represent visual data in a symbolic form by training the network to identify more relations (spatial and others) and objects by designing the network to predict both subject and object labels as well as the relation label. For example, we want to go from an image of a ball above a boy to a representation in RDF triple as (Ball, Above, Boy) or in first order logic as 

\[ \exists{x} \exists{y}\ Ball(x) \land Ball(y) \land Above(x, y) \]

Also, the network can be extended to generate relations for multiple subject-predicate-object combinations in natural images to improve descriptions of the images. We also see the need to create a larger dataset with real world images to train better networks for spatial relations. We plan to use image captioning datasets such as MS COCO \cite{lin2014microsoft} and Flickr30k \cite{plummer2015flickr30k} and annotate the images with spatial relations between object pairs by parsing the image descriptions. We would like to apply our analysis method to such real world dataset.

% sigproc.bib is the name of the Bibliography in this case
\bibliographystyle{abbrv}
\bibliography{sigproc}

\end{document}